\pdfoutput=1

\documentclass[11pt]{article}

\usepackage{ACL2023}

\usepackage{times}
\usepackage{latexsym}

\usepackage[T1]{fontenc}

\usepackage[utf8]{inputenc}

\usepackage{microtype}

\usepackage{inconsolata}

\usepackage{longtable}
\usepackage{setspace}
\usepackage{graphicx}

\usepackage{enumitem}

\title{Dialogue Games for Benchmarking Language\\ Understanding: Motivation, Taxonomy, Strategy}

\author{David Schlangen \\
  Computational Linguistics / Department of Linguistics \\
  University of Potsdam, Germany\\
  \texttt{david.schlangen@uni-potsdam.de}}

\begin{document}
\maketitle
\begin{abstract}
  How does one measure ``ability to understand language''? If it is a person's ability that is being measured, this is a question that almost never poses itself in an unqualified manner: Whatever formal test is applied, it takes place on the background of the person's language use in daily social practice, and what is measured is a specialised variety of language understanding (e.g., of a second language; or of written, technical language).
  Computer programs do not have this background.
  What does that mean for the applicability of formal tests of language understanding?
  I argue that such tests need to be complemented with 
  tests of language use embedded in a practice, to arrive at a more comprehensive evaluation of ``artificial language understanding''.
  To do such tests systematically, I propose to use ``Dialogue Games''---constructed activities that provide a situational embedding for language use. I describe a taxonomy of Dialogue Game types, linked to a model of underlying capabilites that are tested, and thereby giving an argument for the \emph{construct validity} of the test. I close with showing how the internal structure of the taxonomy suggests an ordering from %
  more specialised to more general
  situational language understanding, which potentially can provide some strategic guidance for development in this field.

\end{abstract}

\section{Introduction}

\begin{figure*}[ht]
  \centering
  \includegraphics[width=.7\linewidth]{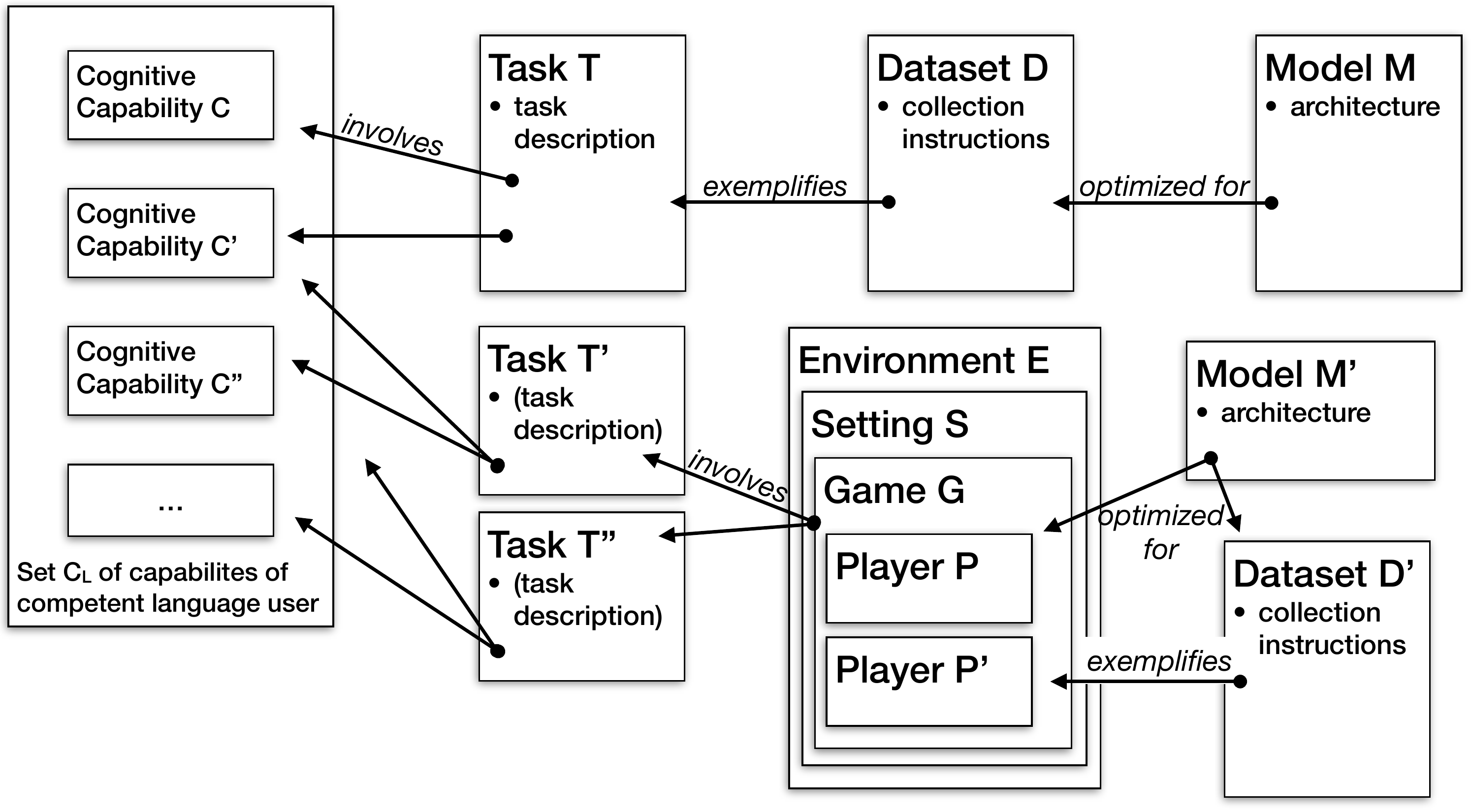}
  \vspace*{-1ex}
  \caption{The structure of relations between the research objects \emph{model, dataset, task, environment, setting, game}, and \emph{cognitive capability}. Adapted from \cite{schlangen:tasks}.}
  \label{fig:tag}
  \vspace*{-2ex}
\end{figure*}

\emph{Steff is sitting at a desk, intently focussed on the piece of paper in front of them. The task is to read short paragraphs of text and then to answer questions about them, and how well Steff does at this will determine their ``language proficiency score'', and thus contribute to whether they will get admission to the University of their choice or not -- for it is a requirement to \emph{understand} the language that is being tested. Steff gets up and heads toward the door -- ``\emph{the other one}'' shushes the proctor. Outside, Steff sees their friend, looking at them, and greets them with ``\emph{next time}''; the reply comes immediately: ``\emph{drinks}?''}

The subfield of ``Natural Language Understanding'' (NLU) within the field of Natural Language Processing (NLP) uses tests of the first kind---written responses to written material---to measure the degree to which a technical artefact can be said to possess the \emph{ability} of understanding natural language. %
More recently, NLP has expanded towards tackling more situated and less abstracted cases of language use---as in the second part of the story, if not quite as social---, under the headings ``language and vision (navigation)'' or ``embodied AI'' \cite{Duan2022,gu-etal-2022-vision,sundar-heck-2022-multimodal},\footnote{%
  The field has moved \emph{back} to this, one should say, as of course situated language used to be much more in the center, as for example in the very early SHRDLU system \cite{winograd:shrdlu}.
}
with evaluation practices not yet fully established.

This paper aims to systematise already ongoing efforts in this direction and to support future ones, by first asking how these kinds of language understanding settings---formal, and situated---relate. Coming to the conclusion that Situated Language Understanding (SLU) requires different testing approaches, and that NLU evaluation has proceeded somewhat haphazardly, I will describe the design choices for creating situated language use activities, relating them to a particular, but abstract, model of situated language understanding; thereby addressing for this new field the concern of \citet{schlangen-2021-targeting} that progress cannot be measured without clarity about underlying theoretical commitments. More specifically,
I want to show a way how Dialogue Games can be integrated into a sound methodology for computational research on meaning, by providing explicit information about relations between research objects (see Figure~\ref{fig:tag}).

It might be useful to mention at the outset what this paper is \emph{not} aiming to do, which is to make recommendations for \emph{how} SLU should be modelled, in the technical sense. While I see value in being able to understand the components of a task and how they interact (which suggests modularity in design), nothing precludes attempting benchmarks of the types described here with monolithic models, and even, insofar as the requisite information can be represented in the right way, with ``general purpose'' models such as Large Language Models.

\section{Background On Measurement}
\label{sec:measuring}

\emph{Language understanding}, as a psychological process, is observable only in its reflections in behaviour.\footnote{%
  This is independent of whether you think that it is a process resulting in a specific psychological state, or a behavioural disposition \cite{Ryle:Concept}.
}
But not any behaviour counts, and not any behaviour is \emph{measurable}---and measurement is our goal here. Experimental psychology has developed many ways to deal with the problem of measurement of unobservables in a principled manner. A central notion here is that of \emph{validity} of a measurement instrument: Does the instrument indeed measure the unobservable \emph{construct} that it is set up to measure?

This is not the place to give a full introduction into that field,\footnote{%
See \cite{franketal:experimentology,Flake2020,Sireci2013} for some recent overviews.
}
so I will concentrate on those aspects of validity that I see as attainable through the methodology described below.
A first claim for validity of an instrument is via an appeal to its \emph{face validity}: That it intuitively appears to capture the construct. Being able to count in a text the occurrences of the letter ``o'', for example, would lack such face validity for the construct ``language understanding'', while being able to answer questions about it may be argued to have it. (Although our intuitions leave us quickly here: What if some questions are answered well, but others bizarrely badly? More on this below.) A second element is \emph{ecological validity}, an argument for how closely the measure resembles the use of the construct in the domains in which it ordinarily shows. Measures of situated language understanding (as will be developed here) can arguably make a claim for high ecological validity---this kind of language understanding plays a large role in people's lives---but more abstract or formalised understanding tasks do occur, in situations as described above. Lastly, quantifiable support comes from \emph{convergent validity}, as different measures that are purportly addressing the same construct can be expected to correlate, and if they do so, mutually support their validity.

What is important to note is that behind all these aspects of validity there is a argumentative connection to the construct and its structure, lending a kind of network character to the notion: ``the measure is valid if there is evidence that it fits into the nomological network – the network of predicted relationships with other constructs and their measures'' \cite{franketal:experimentology}. We will see below that this is something that is missing in the evaluation practices in NLU, and it is something that I will try to develop here for SLU. (In Figure~\ref{fig:tag}, this is the box on the left.)

Summarising this brief review, to avoid ``Questionable Measurement Practices'', \citet{Flake2020} propose a number of questions to which experiment designers must be able to give a good answer \cite[p.\ 459]{Flake2020}:\footnote{%
  These are the first four of the six questions they give; the latter ones concern pre-registration, the use of which in NLP would be a topic for another paper and is glossed over here.
}
\begin{enumerate}[itemsep=2pt,topsep=2pt,parsep=1pt,partopsep=1pt]
\item What is your construct?
\item Why and how did you select your measure?
\item What measure did you use to operationalize the construct?
\item How did you quantify your measure?
\end{enumerate}

\noindent
The questions shall serve as a guide for the discussion below.

\section{Current Practices in Measuring NLU}
\label{sec:nlueval}

The practice of \emph{benchmarking} in NLP / AI is curiously disconnected from that of measurement in experimental psychology, even if it sets itself what looks like rather closely related goals (for example, to provide a ``General Language Understanding Evaluation'', as indicated in the name of the \textsc{glue} corpus, \cite{Wang2019}).

Evaluation in NLU centers on the \emph{language task}, a functional mapping between input and output, where at least one of these involves language.\footnote{%
  The discussion in this section follows \citet{schlangen-2021-targeting}, which however did not yet use the language of measurement from experimental psychology, however; this connection is helpfully made in \cite{Raji-et-al-everything}.
}
For a given NLU evaluation corpus, this mapping is typically characterised verbally; e.g., ``the text labelled `answer' is a correct answer to the question in the text labelled `question', given the context in the text labelled  `passage' '', as this description could go for the example in Figure~\ref{fig:glue}.
It is this verbal (or \emph{intensional}) description that enters into an intuitive appeal to face validity---surely, answering questions must require understanding them. However, it can be remarked that the notion of understanding in NLU evaluation typically remains an intuitive one and no further attempt is made at specifying the construct.

In any case, the actual measurement instrument is one step further removed, as the task needs to be operationalised via \emph{instances} collected into a dataset; this then serves as the \emph{extensional} definition of the task. As observed in \cite{schlangen-2021-targeting}, to not lose connection to the validity argument (which goes via the intensional description) requires care in setting up the dataset, which sometimes is missing. (For example if the collected instances do not span the domain in the way claimed by the intensional description.) For specialised machine learning models, a further challenge is posed by the fact that they are typically trained on a (set aside) portion of the dataset. Machine learning methods are very good at identifying predictors that optimise performance, regardless of whether these predictors are related to the construct that is to be measured \cite{Lapuschkin2019}.   (Note again that for humans, tests of language understanding happen on the tacit and unquestioned background of already existing general language competence, acquired through material distinct from the testing material.)

\begin{figure}[t]
  \centering
  \small
  \fbox{
  \begin{minipage}{.9\linewidth}
\textbf{Passage:} Barq's – Barq's is an American soft drink. Its brand ofroot beer is notable for having caffeine. Barq's, created by Edward Barq and bottled since the turn ofthe 20th century, is owned by the Barq family but bottled by the Coca-Cola Company. It was known as Barq's Famous Olde Tyme Root Beer until 2012.\\
\textbf{Question:} is barq’s root beer a pepsi product\\
\textbf{Answer:} No
\end{minipage}
}
    \caption{An Example of a GLUE-type task (from the BoolQ subset, \cite{clark-etal-2019-boolq}, as cited in \cite{superGLUE})}
  \label{fig:glue}
\end{figure}

With respect to these concerns, practices in NLU evaluation have not much improved. With existing tests saturating when probing newer models, the response has become to go bigger, and efforts such as BigBench \cite{bigbench2022} and HELM  \cite{helm2022} invested in bringing in very many different evaluation sets. While this may be seen as potentially improving convergent validity (if a model achieves high performance on so many tests, it must be doing \emph{something} right), there is still little concern about what exactly the underlying construct is. This we can then take with us to the next section: NLU evaluation centers on language tasks, and relies on the face validity of the task, without making much effort to connect to any further specified construct.

It is also worth noting the criticism of this approach put forward by \citet{Raji-et-al-everything}, which is that the aim of measuring understanding performance in the abstract (without tasks that have extrinsic value beyond their role in the test) through datasets is misguided, conflating as it does a language ability with a recall test on the necessarily open-ended world knowledge that enters into many of these tasks. We will see below how the methodology developed here can answer this reservation, through controlling the world knowledge required to perform. First, we shall look more closely at how NLU and SLU differ.

\section{SLU is Different From NLU}
\label{sec:diff}

To give us more examples of situated language use, %
here is another short story:

\begin{quote}
  \emph{
    You are assembling flat-packed furniture, with the help of your friendly household robot.
  You send the robot to ``fetch the box cutter from the drawer in the other room.''$^{(1)}$ ``Which one, it's not in the one with the tools''$^{(2)}$, you hear it shout from the other room. Later, the both of you look over the instructions -- why are the pictograms always so obscure? -- and discuss how to proceed. Having reached step 24, you look at a screw and wonder whether it is of type 35784, of which there were supposed to be 12 in package A, but the robot just says, ``no, the other one''$^{(3)}$. ``Alright, so can you pass me the torx?''$^{(4)}$, you say. ``Sure, here you go. That's a torx then?''$^{(5)}$
}
\end{quote}

This---obviously constructed, but nevertheless hopefully coherent---story showcases several features of situated language use unlikely to be found in monological text corpora: Example~(1) contains referring expressions that express an \emph{exophoric reference}---a reference to singular objects outside of the discourse itself, but to its immediate situational context---, and it realises a \emph{request} speech act, for which one sign of understanding is compliance through (non-verbal) action. (2)~realises a \emph{clarification request}, which is another way understanding can be signalled, albeit a partial understanding only. This hints at the processual nature of understanding in interaction, different from the single-shot framing as in the example in Figure~\ref{fig:glue}. (3)~highlights how the syntax of situated language can be different from the edited written language found in NLU corpora; it is a ``\emph{non-sentential}'' or ``fragmental'' utterance of a kind which is frequent in dialogue and not at all syntactically or semantically malformed \cite{schlasc:edilog,fernginz:corp}. It also shows that the acts that are to be understood need not be linguistic ones; here, the fragment itself is a reaction to a presumed mental state.
(4)~again is a request for action, this time in the guise of a question as an \emph{indirect speech act}. (5)~finally shows that an outcome of understanding in situated interaction can be that understanding itself can be \emph{adapted}---here, we would expect an agent that has real understanding to be able to later use the term that at that point was new to them.

What this short example has shown, when contrasted with the example in Figure~\ref{fig:glue}, is that SLU differs both on the side of the ``input'' (the act that is to understood) as well as on the side of the ``output'', where the action space is much larger---in fact, infinite (but compositional).
That is, SLU poses language tasks that do not occur in the text corpora used in NLU research.
Even more importantly, the individual acts of understanding (from one turn to the next) are embedded in the general goal-directed structured of the interaction as a whole; something that cannot be captured in the i.i.d.\ (independent and identically distributed) nature of a static dataset. This argues for finding a measurement instrument that provides not only richer context information in a static way (as could be recorded in a richer dataset), but also an active embedding of language use in varying, goal-directed interactions.

\section{SLU Requires Different Benchmarking Methods: Dialogue Games}
\label{sec:dg}

The scenario described above makes for a good use case---having such a robot would be useful!---but a bad measurement instrument. One reason for that is that it simply is far out of reach of current technology (not just in the language abilities, but also in the physical abilities that it suggests). Any attempts at approximating such abilities with current technologies would require making design choices that are more driven by the specific goal rather than by testing language abilities. This also suggests a second problem, which is that this scenario does not isolate the language abilities well enough to serve as a good test. We hence need more controlled situations in which the situated language use can be modelled, while preserving the goal-orientation exhibited by this scenario, as the structure it provides is, as argued above, a crucial element that is not captured by dataset-based methods.\footnote{%
  This can be seen as a restriction compared to the general phenomenon of Situated Language Use: Not every language use situation must necessarily be understood as goal-directed. However, if interaction episodes generally are seen as having a beginning and an ending \cite{clark:ul} and the notion of activity types \cite{Levinson1979} is accepted, a broad goal of getting from beginning to ending can be assumed to be active in general.
}

This discussion motivates the use of what I will call \emph{Dialogue Games} as benchmarking instrument, where:\footnote{%
  Named of course with a nod to Wittgenstein's \emph{language games}: ``I shall call the whole, consisting of language and the activities into which it is woven, a `language game''' \cite[§7]{Witt:PU-corr}; with another inspiration coming from Levinson's ``activity types'' \citep{Levinson1979}.
}$^,$\footnote{%
  Note that this definition is general enough to cover ``book a train ticket'' or even ``interactively instruct agent to summarise a text'' under the name ``game'' as well.
}

\begin{quote}
  A \emph{Dialogue Game} is a constructed activity with a clear beginning an end, in which \emph{players} attempt to reach a predetermined \emph{goal state} primarily by means of producing and understanding linguistic material.
\end{quote}

\noindent
It is the goal-orientation and the constructed nature of the activity, as we will see, that makes it possible to target particular aspects of Situated Language Understanding, without conflating understanding with recall performance on general world knowledge.

Before moving on to how such games can be constructed in such a way that they make clear connections to (assumptions about) underlying capabilities, and how they can be organised into a strategic plan for making progress, we need to register a cautionary note from the (long) history of this type of approach.
In 1972, Minsky and Papert
introduced the notion of ``micro-world'', as a way to explore ``intelligence'' problems in context: ``we see solving a problem often as getting to know one's way around a `micro-world' in which the problem exists'' \citep{MinskyPapert1972}. The most famous of these micro-worlds is the ``blocks-world'' of the system SHRDLU \cite{winograd:shrdlu}---which would count as a Dialogue Game according to the definition above. SHRDLU seemed to demonstrate what I have called here Situated Language Understanding quite well, but criticism of the approach soon arose, of which the following quote is representative (see also \citet{Dreyfus1981,Marr:Vision}):

{\small
  \begin{quote}
    \it
  SHRDLU performs so glibly only because his domain has been stripped of anything that could ever require genuine wit or understanding. [\dots] Neglecting the tangled intricacies of everyday life while pursuing a theory of common sense is not like ignoring friction while pursuing the laws of motion; it's like throwing the baby out with the bathwater. A round frictionless wheel is a good approximation of a real wheel because the deviations are comparatively small and theoretically localized; the blocks-world ``approximates'' a playroom more as a paper plane approximates a duck.
  \cite[p.\ 190]{Haugeland1985}
\end{quote}
}

This gives us a warning to bring with us to the further discussion, which is to take care that any abstractions made in simulation shall not abstract away the real challenges.
We will come to a delineation of the design space in which we can search for Games that meet this challenge in a moment, but first I will sketch a model of the capabilties underlying SLU, to which we can then connect the Game taxonomy.

\section{The Construct: A Model of Capabilities Involved in SLU}
\label{sec:modcap}

\begin{figure}[t]
  \centering
  \hspace*{-3ex}
  \includegraphics[width=1\linewidth]{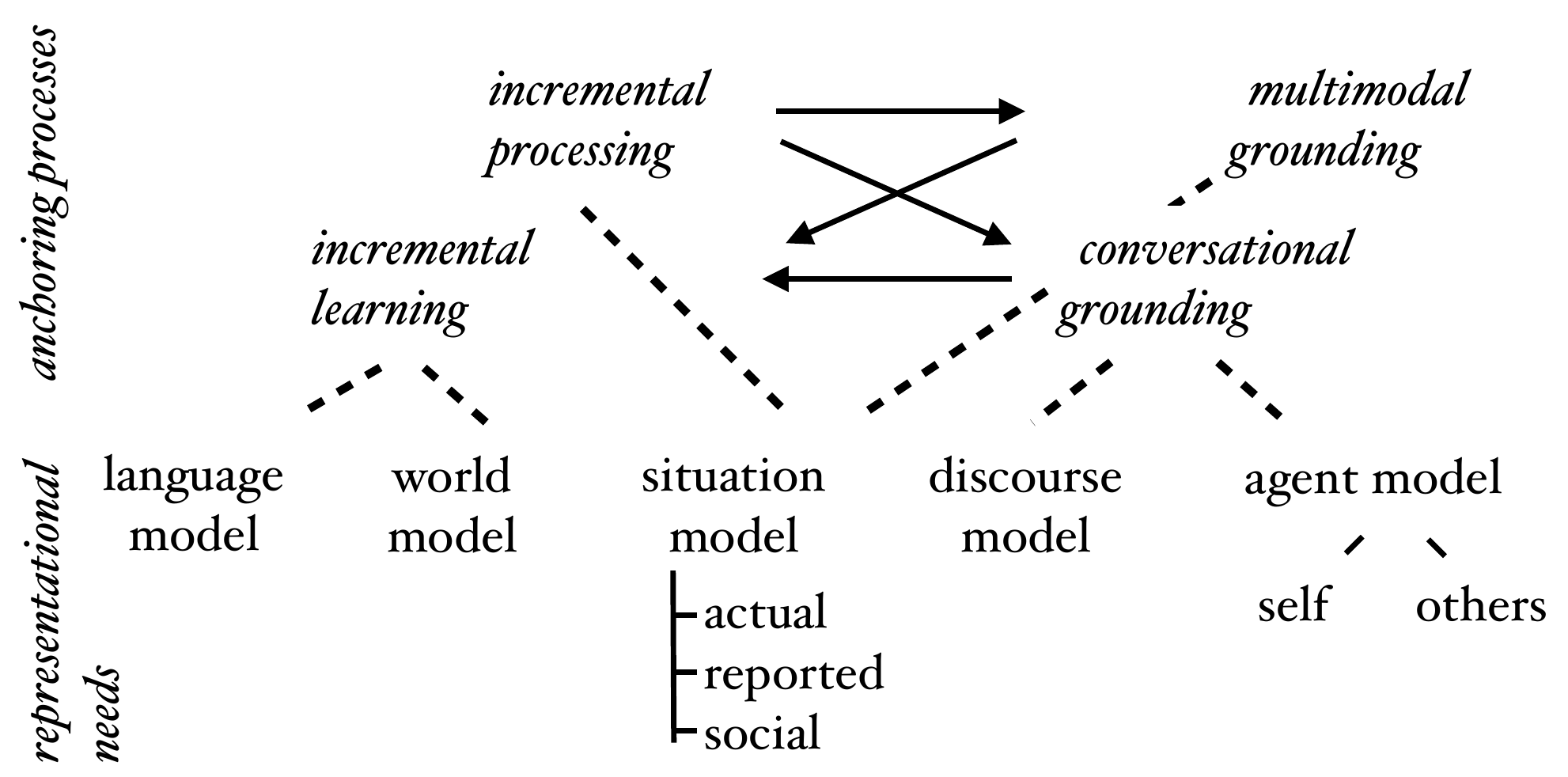}
  \caption{Representational Domains (bottom) and Anchoring Processes (top) Structuring the Situated Agent}
  \label{fig:analysis}
  \vspace*{-2.5ex}
\end{figure}

The methodology described above (illustrated in Figure~\ref{fig:tag}) rests on the benchmarking instruments being explicitly grounded in (assumed) capabilities that are being tested. Elsewhere, I have developed a model that distinguishes between various kinds of capabilities involved in SLU \cite{Schlangen-2023}; this will serve us here as the ``nomological network of relationships between constructs'' from Section~\ref{sec:measuring} above.

This model, illustrated in Figure~\ref{fig:analysis}, assumes that the agent represents what I call ``knowledge domains'', and maintains ``anchoring processes'' that operate on them. The knowledge domains are as follows: the \emph{language model} (here meant to collate only linguistic knowledge about the form/meaning mapping; updated rarely), the \emph{world model} (concepts, concept hierarchies, script knowledge, etc.; also updated rarely), the \emph{situation model} (details of the current conversational situation and/or the reported situation; updated continuously), the \emph{discourse model} (what has been said so far, and how it relates; discourse referents; updated continuously), and finally the \emph{agent model} (of the beliefs, desires, intentions of agents, and recursively what it represents of the participating agents; also updated continuously).

\begin{figure*}[h!t]
  \centering
  \includegraphics[width=.7\linewidth]{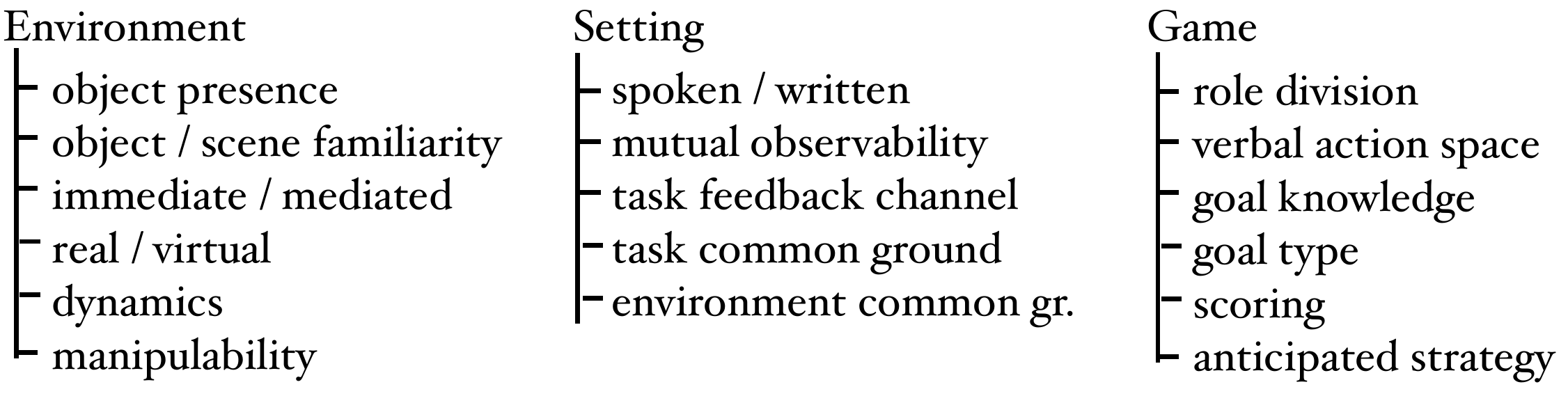}
  \vspace*{-1ex}
  \caption{The main components of the proposed taxonomy}
  \label{fig:tax}
  \vspace*{-2ex}
\end{figure*}

As anchoring processes (which ``bind the agent to the here, now, and us''), there is \emph{incremental processing} (updating situation and agent model, based on minimal units of observation), \emph{conversational grounding} (the process of negotiating shared understanding, for example through asking for clarification, if necessary), \emph{incremental learning} (which ranges from the establishment of spontaneous local conventions, e.g.\ on how to refer to objects, established during the conversational grounding, to learning facts from observation and from testimony, and, crucially, from discussion and disagreement), and \emph{multimodal grounding} (resolving references to objects in the shared surrounding, as well as deriving meaning from non-verbal actions such as gestures).\footnote{%
  For a more detailed description and a justification of this way of analysing the SLU agent, including references to prior work making use of releated concepts, see the original paper \cite{Schlangen-2023}.
  }

  What this gives us is a finer-grained picture of the construct: \emph{understanding} means applying, building up and maintaining these representations, via these processes.\footnote{%
    Let me stress again that I am not making any claims about whether such representations should be built into models of situated language understanders or not; the (falsifiable) claim is just that something like them will be found in such models.
  }
Not all acts of understanding rely on all aspects equally, and this makes it possible to develop a strategy for working towards modelling the overall capability, as we will see. With this in hand, we can now come to the design aspects of Dialogue Games, and how they may put certain of these capabilities (knowledge domains and anchoring processes) more or less into focus. You can read the following section also as advice on best practices in experiment design, drawn from extensive experience in setting up what is here systematised as Dialogue Game \cite{deawu:pttarub,kenningtonetal:sigdial13,pentoref2016,ilieatal:meetup19,attari:sigdial19,Schlangen-2019-2}.

\section{A Taxonomy of Game Types}
\label{sec:tax}

The most salient aspect of a Dialogue Game might be the task that it poses to the players; that is, the goal state and how to get there.
For what could be called the ``furniture assembly game'' from Section~\ref{sec:diff}, this would be 
\emph{have the furniture fully assembled} (goal state) and \emph{provide required assistance} (game ``play''); for a more realistic Dialogue Game (of the type ``reference game'', see below), that could be \emph{bring cards into same order} (goal state)  and \emph{ask each other which cards there are, and jointly decide on order} (game play).
But hidden behind descriptions like these there is a large number of additional design decisions that need to be made before the game can be played. These decisions have many degrees of freedom, but all come with subtle influences on the shape of the interaction, and on the phenomena that one can expect to see in protocols of the game play.

I distinguish here between three major areas in which decisions must be made when setting up a concrete Dialogue Game: Enviroment, Setting, Game Proper; with many sub-aspects within. A comprehensive overview is given in Appendix~\ref{sec:A}. The discussion here presents these design features and directly links them to the model of the underlying construct(s) from the previous section.

\subsection{Environment}
\label{sec:env}

This section groups together all design decisions that influence \textbf{what the relevant entities and actions in the game are}, and \textbf{how they are presented to the players}.

A high-level decision here is whether the game requires talking about \textbf{objects that are currently present} (in some form), or not. (An example of a task that is not about currently present objects would be booking a train ticket, which does require talking about entities such as train stations, without them needing to be present. Such a conversation is still situated in the sense that the interlocutors share time, but it is at the boundaries of what I consider here.) This decision influences how the \emph{situation model} is constructed (e.g., from visual evidence or not) and how the \emph{world model} is challenged (because for non-present objects, agreement on referents must come from prior common ground).
A different design dimension concerns \textbf{prior knowledge of these objects}, whether they are (expected to be) \textbf{familiar} to the players or not. Whether something is familiar or not depends on the \emph{world model}, and whether it is assumed to be \emph{mutually} familiar on the \emph{agent model}; succesfully referring to unfamliar objects means more effort in \emph{conversational grounding}.

Another decision is whether \textbf{access} of the players to the objects is \textbf{immediate or mediated}, and if mediated, if the \textbf{objects are real or computer simulated}. (So, a video call would be mediated but real; operating with representations on a computer screen would be mediated and virtual.)
Virtual environments make further abstractions possible, for example by discretising changes (the world ``jumps'' from one state to the next), or reducing the action space (what can be done to and with objects).
The difference here is less one in what capabilities are challenged than in the control that is given over the situation; for example, a \textbf{static} environment will force fewer updates to the \emph{situation model} as one with \textbf{discrete} updates, which in turn may require slower changes than one that is fully \textbf{dynamic}.

\subsection{Setting}
\label{sec:set}

This dimension collects decisions about \textbf{how the players can interact with each other}---which of course determines to a large extent what kind of data can be expected.

A first high-level decision here is whether the verbal interaction is done via \textbf{speech}, or through \textbf{typed messages}. Written language, even in the dynamic form that it can take in chat interactions, is a restricted channel compared to spoken language (where prosody and other para-linguistic information provides a channel for \emph{multimodal grounding}); the interaction also slows down and is, at least in typical setups, more discretised (as messages need to be sent before they are seen; this influences the degree to which \emph{incremental processing} is challenged). Finally, turn-taking, which is an essential process in the organisation of free interaction \cite{deruiteretal:corner}, works differently in chat communication than in spoken interaction. On the other hand, practical advantages in choosing typed messages are also clear, in that written language is typically easier to store, and (for artificial agents) to process, and to generate.

Another set of decisions concern \textbf{what the players see of each other}, and \textbf{what they see of their actions} in the environment. \emph{Multimodal grounding} of the signal in actions of the interlocutor is an important aspect of meaning making \cite{Holler2019}; disabling it through hiding the interloctor forces more meaning into the verbal channel (which can be desired, but reduces ecological validity). Similarly, other aspects of interaction management get harder when there is no visual contact between interloctors \cite{brennan:acl00}, but at same time become more visible in the linguistic material. 

I have also grouped under this heading questions of how \textbf{common ground} between the participants (other than what they can see of each other) can form. If the players knowingly play repeated rounds of the game (from some initial state to a respective goal state), they can build up personal common ground \cite{clark:ul}, a form of \emph{incremental learning} influencing their \emph{agent model}.\footnote{%
  These days in Artificial Intelligence more typically called ``theory of mind'', see e.g.\ \cite{bara-etal-2021-mindcraft}.
} When players (knowingly) \textbf{share the same environment} (be that a simulated and mediated one or a real one; where even looking at the same image would count as sharing the environment), there is an automatic assumption that large parts of the respective \emph{situation models} are shared (and represented as thus in the \emph{agent model}); if this is not the case, or not knowingly so, linguistic labor must be performed to reach such common ground (if the tasks requires it).

\subsection{Game}
\label{sec:game}

The decisions grouped here concern the game in the narrow sense: \textbf{how initial state and goal state are defined}, but also \textbf{what the players know about this}, and \textbf{how the games defines roles} and \textbf{suggests strategies}. In the terminology of \citet{suits:grasshopper}, a game must subordinate under a \emph{prelusory goal}, which can be stated independently of the game (e.g., in football (soccer), ``make the ball be in the opponent team's goal''); it is further defined by \emph{constitutive rules}, which make reaching that goal more difficult than necessary (e.g., by disallowing to just grab the ball and carry it to the goal); it must also trigger in the players a \emph{lusory attitude}, which is the acceptance of the complications posed by the constitutive rules. Doing the latter successfully can increase the quality of the collected data, as players with higher engagement can be expected to show a wider range of behaviours \cite{vonAhn:espgame}. Inspiration can be taken here from the literature on the design of games (e.g., \cite{adamsdormans:gamemech}); ultimately, however, the purpose of a Dialogue Game in the sense developed here is to provide data and a testing environment in a principled way, and not primarily enjoyment.

To classify games, one high-level aspect concerns the \textbf{goal type}, where we can distinguish \textbf{reference games} (a time honoured instrument in Psycholinguistics, going back at least to \cite{krausswein:1964}; see \cite{ji-etal-2022-abstract} for a recent overview), which focus on reference and hence \emph{multimodal grounding} and alignment of \emph{situation models}; \textbf{information games}, which center on the requesting and giving of information, which depending on the domain can lead to demands on the \emph{world model} and/or the \emph{situation model}; \textbf{construction games}, which go beyond reference and information in that they require the execution of actions, and hence require coordination of the anchoring processes to a higher degree; \textbf{navigation games}, which center spatial language and spatially complex \emph{situation models}; \textbf{negotiation games}, which focus on explicit coordination of \emph{agent models}; and finally \textbf{teaching games}, which make explicit the \emph{incremental learning} and how it updates the \emph{world model}. This is not a complete categorisation, and each concrete game will contain elements of more than one of these types; but this does represent good coverage of types of games typically used. (We are currently preparing a comprehensive survey of the field, which will provide a plethora of references for representatives of all types.)

Some more final subdimensions. In the design of the game, the player can be assigned \textbf{distinct roles} with different responsibilities, such as for example an assigned \emph{questioner} paired with an assigned \emph{answerer}, or an \emph{instruction giver} with an \emph{instruction follower}. The stricter these roles are, the lower the coordination effort required, deemphasising functions such as \emph{conversational grounding} and the keeping of detailed \emph{agent models}. Goal-directed games naturally come with a notion of \textbf{success}, but beyond that, \textbf{scoring} functions can be introduced (for example, ``faster is better''; or a reconstruction loss for construction-instruction tasks). Making the score known to players introduces incentives that can change the dynamics of the interaction (e.g., proritizing speed over accuracy, or vice versa). 

Finally, the design of the game can also make a desired \textbf{strategic behaviour} more salient. A \textbf{cooperative} player would be one who does their best in understanding intents behind requests (e.g., through replying correctly to indirect speech acts, or to providing partial information when a question cannot be answered fully), whereas a \textbf{collaborative} player is one who takes their goal to be shared with the other player, and who hence has an interest in being proactive as well---likely to be more challenging to the anchoring processes and to the alignment of the \emph{agent model}s.\footnote{%
  We note here that a demand for cooperation can lead to increased coordination effort, which can result in the players negotiating in the game to follow a merely \emph{cooperative} strategy (with one instruction giver and one instruction follower), if this seems more efficient to them. %
  This is something that we have experienced with the MeetUp game \cite{ilieatal:meetup19}, where two players moving in separate copies of the same virtual environment must manage to meet up in the same room (without seeing each other), and which we had designed to trigger collaborative interactions; it however turned out to be a frequent strategy for one player to just stop moving and only answer questions by the other.
}

\section{Dialogue Games as Evaluation Instrument}
\label{sec:eval}

Let us assume that we now have designed a Dialogue Game, starting from ideas about which aspects of the construct we particularly want to challenge and making careful decisions on all the design features mentioned above. Of the questions listed above in Section~\ref{sec:measuring}, we have an answer to numbers 1 to 3 (Q1: What is the construct?---A: The model in Section~\ref{sec:modcap}); Q2,3: Why and how did we select measure? What measure to operationalise?---A: By making a decision to focus on some aspects, and selecting according to Section~\ref{sec:tax}). This leaves one crucial element, Q4: Deciding on how to quantify the measure. This for us translates into how to use the dialogue game to quantify the abilities of an artificial model, which is what this section will look into.

Dialogue games can be used as a means for data generation, simply by collecting game play from people playing the game. Using a dialogue game promises to offer some control over the data that is to be expected, insofar as the connection between properties of the game to underlying capabilities (as discussed in the previous section) is also reflected in properties of the language use. For example, a reference game will make the use of referring expressions likely; a game where mutual understanding is particularly forwarded will make linguistic devices for conversational grounding prominent \cite{Schlangen-2019-2}. This can be interesting for the study of these linguistic phenomena (see, e.g., \cite{deawu:pttarub,schlandez:is07}). Moreover, the control over environment and setting makes it possible to record rich contextual information alongside with the language use \cite{kousidis:mintlab}.

This does not mean, however, that the use ends with the recording of richer corpora, to then be used in the same way as the NLU corpora mentioned above. In fact, as we have touched on above, using static corpora for research on SLU is problematic, as here the recorded actions can count even less as reference than they do in other generative tasks, such as machine translation.\footnote{%
  Where metrics that compare model predictions against a reference have long been criticised, see \emph{inter alia} \cite{turian-etal-2003-evaluation}; see \cite{liu-etal-2016-evaluate} for an early extension of this critique to the evaluation of ``dialogue systems''.
}

In this section, I will highlight some uses that go beyond the train/val/test dataset paradigm typical of NLP, all revolving around the ability of Dialogue Games---at least those with simulated environments---to serve as \emph{execution environments}.

\paragraph{Rollout}

One such use, which in a way stands between the use of static corpora and the evaluation of agents in interactive game play, is nicely exemplified by the benchmarks built on the TEACh dataset \cite{Padmakumar2022}. The game, according to our taxonomy, uses a simulated, dynamic environment with familiar objects (household objects), is a Navigation Game with elements of an Information Game (as a Commander, for whom the environment is fully observable, instructs via a written channel a Follower, for whom the environment is only partially observable, to perform a task in the environment, with the Follower getting an opportunity to ask for clarification). Two tasks are defined that one may call \emph{rollout} tasks (our terminology), in the sense that they require the prediction of actions, based on partial or complete history. Crucially, since the environment simulator is provided, the predicted actions can be executed, and the evaluation target is whether the required state changes have been affected. This abstracts away at least to a certain extent from what is recorded in the corpus, as only the state changes (and not all actions) serve as reference. This type of evaluation is only possible in a Dialogue Game setup and not with a static corpus alone.

\paragraph{Agent/Agent Play}

If artificial agents for all roles in the game are provided, another mode of evaluation comes available, that of fully simulated play. Games as defined above will come with a some sort of score that measures success in reaching the pre-defined goal state, and this can then serve as the evaluation target. If a human/human reference corpus is available, the produced language itself can then furthermore be evaluated along formal parameters (e.g., average turn length and distribution, vocabulary size, etc.).

Note however that the agent/agent mode leads to a even further deviation from the ``test is like training'' approach described above for NLU, as, unlike in non-communicative game like arcade-type games---one of the early successes of new-generation reinforcement learning, e.g.\ \cite{DBLP:journals/corr/MnihKSGAWR13}---this setup can here not be used for learning the agents: In the language case, a competent player needs to already exist from which competent language use can be learned.\footnote{%
  A problem that has led to the research area of ``language emergence'' in what could be called Dialogue Games between deep reinforcement agents, where however the agents are allowed to coordinate on their own language system, folding the language evolution and language learning problem into one. (See e.g., \cite{DBLP:conf/iclr/LazaridouPB17}.) We concentrate here on the setting of evaluating agents that have acquired (to the extent needed for the Game) an existing natural language.
}

We note that Dialogue Games (in our sense, as combination of environment, setting, and game) provide an interesting perspective for the evaluation of (supposedly) general-purpose ``foundation models'' \cite{Bommasani2021FoundationModels}, which have been claimed to be able to function as general simulators of agents \cite{andreas-2022-language}.

\begin{figure*}[ht!]
  \centering
  \includegraphics[width=.7\linewidth]{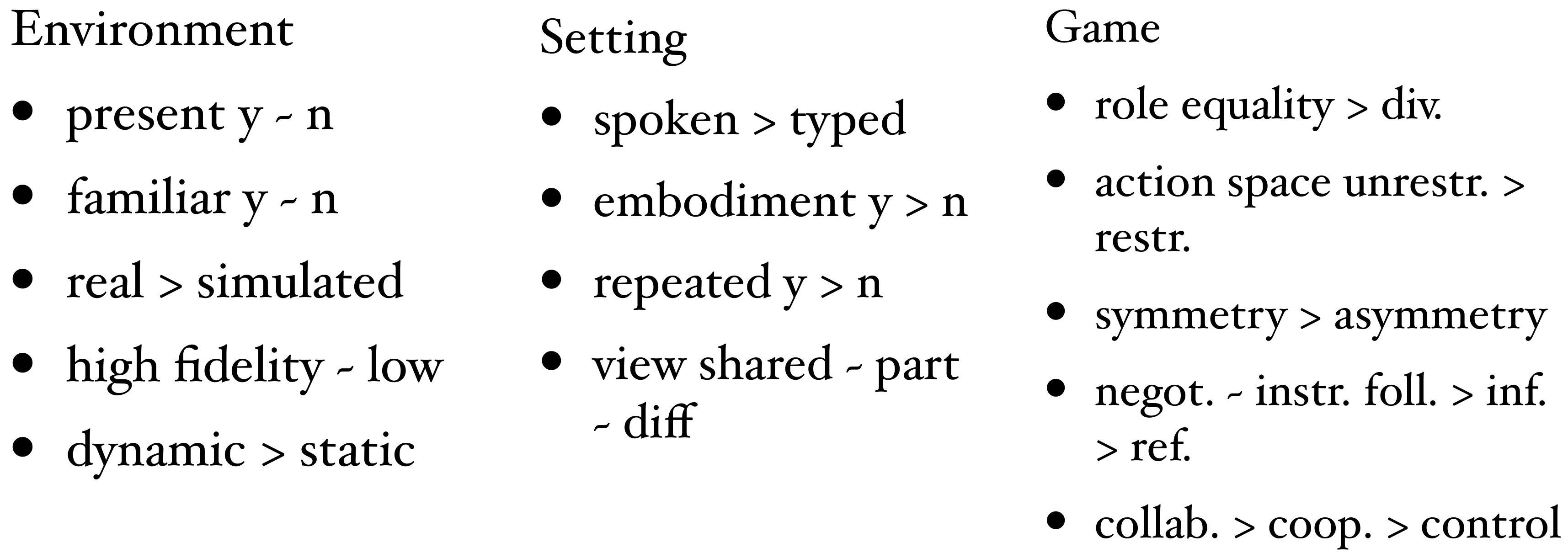}
  \vspace*{-1ex}
  \caption{A partial order on the space of Dialogue Games. $\sim$ denotes ``similar complexity'', $>$ denotes ``leading to higher complexity''.}
  \label{fig:ordering}
  \vspace*{-2ex}
\end{figure*}

\paragraph{Human/Agent Play}

The most informative evaluation mode of interactive system remains evaluation in actual online interaction with human players.\footnote{%
  Compare to what is called ``human evaluation'' in many fields of NLP (see \cite{howcroft-etal-2020-twenty} for recent critical overview of human evaluation practices in Natural Language Generation).
}
Beyond the measures defined by the game (measuring task parameters) and other such measures, which 
in the dialogue systems community usually are called \emph{objective measures} such as dialogue length, etc. (see \cite{walkeretal:paradise} for a seminal reference), this mode also makes it possible to evaluate the interactive \emph{experience}, through \emph{phenomenological} or \emph{subjective measures} elicited in questionnaires \cite{10.1093/iwc/iwz015}.\footnote{%
  Although not many fully validated scales exist; a popular one is Goodspeed questionnaire from the neighbouring field of Human/Robot interaction \cite{Bartneck2009}.
}

One possible objection against this evaluation mode is quickly dismissed: While there may be a superficial similarity to Turing's imitation game \cite{turing:test}---what is now known as the ``Turing Test''---in that quality of a conversation is an evaluation criterion, it is important to note that \emph{deception} about whether a player is human or machine need not be, and in most cases is not, part of the evaluation; and this is what is commonly criticised (see e.g., \cite{Levesque2014}). To keep clear of the deception incentive, however, it is important that such subjective measures do not on their own become optimization targets and are always combined with objective, task-oriented measures \cite{edlundetal:humlike}.

\paragraph{Representation Probing}

Another way to gain information about a game-playing model is via \emph{representation probing}.
To cite just one example, \citet{Madureira-2022} use existing dialogue models as what could be called ``overhearers'' of dialogues from a corpus, and probed whether they, at a given state of the overheard dialogue, represented information about the assumed common ground status of propositions. Techniques such as these can help further validate claims about the link between capabilities and games (if the assumption is that the game challenges a certain capability, we would expect to find information that it is based on to be represented during the model processing).

\vspace*{\baselineskip}
\noindent
Let us take stock. Our long journey has taken us from an argument that Situated Language Understanding needs to be evaluated in the context of, well, situated interactions, through the claim that the underlying construct is multi-faceted, to a recipy for constructing measurement instruments---Dialogue Games. For a given Dialogue Game, the recipy provides at least the beginnings of a validity argument, through the links between taxonomy and construct. We also now know that there is a variety of ways of making use of the instrument and deriving from it a quantified measure, which in turn facilitaties (with the usual caveats) a comparison between models. (``Bigger is better.'')
The introduction promised more, however, namely the derivation from this of a general strategy that might get us, eventually, from simple games to scenarios like the robo-helper described above. This is what the next section will discuss.

\section{Strategy: From Simple to Complex Dialogue Games}
\label{sec:strat}

In this paper, I have tried to use insights from assessment in experimental psychology to suggest improvements in practices in assessment in Artificial Intelligence. There is a point, however, at which the similarities end. AI, as a constructive discipline, aims to \emph{build} the artefacts it studies; experimental psychology aims to understand existing (biological) systems. I specifically made the point in the introduction %
that situated interactions of the kinds discussed here come easy to most people; as assessments of understanding abilities for people, Dialogue Games would not have much purchase.\footnote{%
  Perhaps more so when players are restricted to what for them is not their first language; but we leave this unexplored here.
} 
On the other hand, if we look at the current state of the field, it is clear that we are still at the relatively low end of game complexity. To pick two examples, ``visual dialogue'' \cite{dasetal:visdialrl} represents perhaps one of the simplest game types: player A sees an image and a caption, player B only the caption; player B asks 10 questions about the image---without any further goal---which player A must answer). The TEACh benchmark that was mentioned above \cite{Padmakumar2022} still relies on a relatively simple game (the tasks are things like ``fetch a potato'' or ``put all plates on the table''), and still, the performance of even the best models is quite modest. The historical development that has lead to these games is easy to reconstruct and leads to them from non-interactive tasks (image captioning, natural language navigation), extended to the ``adjacent possible''.

The taxonomy described here offers a view towards what is not only adjacent and possible, but also ``uphill''. To make a given game more complex, a simple parameter is to increase the variability of the entities that it involves. Then, restrictions can be removed successively, e.g.\ by going from a static to a dynamic environment, from typed to spoken interaction, and so on. (Figure~\ref{fig:ordering} suggests a partial complexity order by ranking possible feature values.) Following the discussion above, many of these changes will also shift or extend the way the game challenges understanding, and a model capable of this change thus shows itself to reach a higher level. In this way, the taxonomy also suggests a way to construct a meta-benchmark on which (hypothetical) general models of SLU can be measured. (Among two models that perform comparable on one game, that one will rank higher that performs better on a more complex task.)

\section{Related Work}
\label{sec:relwo}

What I call Dialogue Games here has been used for a long time as instrument in driving forward research on situated language modelling. This is not the contribution of this paper---there is a rich, and ever more strongly growing, literature making use of such games (see \cite{Duan2022,gu-etal-2022-vision,sundar-heck-2022-multimodal}; and our forthcoming general survey). What there is less work on is on this instrument itself. \citet{Bisk2020} make general points about types of information the availability of which during learning might improve the ``understanding'' of AI models. \citet{DBLP:journals/corr/abs-2211-08371} cover similar ground to this paper, but more generally focus on the kinds of contexts needed for certain pragmatic phenomena.

There are now several simulation environments that support setting up Dialogue Games in simulated, continuous environments with high fidelity \cite{gu-etal-2022-vision}. In my research group, we have focussed on the development of a flexible environment that makes the implementation of Dialogue Games and the collection of game play for example through crowd sourcing easier
\cite{Goetze-2022-1,slurk.semdial18}.

For a related argument for how SLU differs from a monological perspective, taking a wider Cognitive Science perspective, see \cite{beyond-single}.

\section{Conclusions}
\label{sec:conc}

After briefly reviewing how experimental psychology thinks about the validity of assessments, I reviewed practices of evaluation in NLU, in the light of these considerations of validty. This served as the foil on which to develop a guide for evaluating what I call \emph{Situated Language Understanding} (SLU). I argued that SLU is different from NLU (Section~\ref{sec:diff}) and hence requires different evaluation instruments: \emph{Dialogue Games} (Section~\ref{sec:dg}). As an element in the argument for the validity of this instrument, I reviewed a model of the capabilities involved in SLU, that is, of the internal structure of the construct that is to be measured (Section~\ref{sec:modcap}). This was then followed by the description of a detailed taxonomy of Dialogue Game \emph{types} (Section~\ref{sec:tax}), which can be read as a guide for constructing a particular game in such a way that it can serve as an assessment instrument focussing on particular aspects of the construct. Different ways of using a given game for evaluation were then discussed in Section~\ref{sec:eval}, before Section~\ref{sec:strat} brought together these insights into a discussion of how this suggests a ordering of assessments from simpler to more complex, thereby suggesting a possible development strategy for models in this space.

\section*{Ethics Statement}

Let us address the ethical elephant in the room. Should we even attempt to build systems that can do this kind of situated language understanding? Should research be conducted on increasing the complexity of the tasks in which they can be used, as indicated in Section~\ref{sec:strat}? It should be clear that there are potential enormously beneficial use cases, for example where such systems are used to restore the physical reach of humans that have lost abilities (or never had them). But understanding of the kind discussed here shows in action, making systematic failures or biases of such systems potentially more directly harmful than for language-only systems: What holds for language models holds even more for models with arms. The hope is that the methodology described here of starting with simpler settings \emph{and thoroughly evaluating performance on them} before moving on might provide one way in which this research can be made safer---although of course further work is needed on working out whether this argument holds water.

\bibliography{/Users/das/work/projects/MyDocuments/BibTeX/all-lit.bib}

\begin{thebibliography}{65}
\expandafter\ifx\csname natexlab\endcsname\relax\def\natexlab#1{#1}\fi

\bibitem[{Adams and Dormans(2012)}]{adamsdormans:gamemech}
Ernest Adams and Joris Dormans. 2012.
\newblock \emph{Game Mechanics: Advanced Game Design}.
\newblock New Riders Games.

\bibitem[{Andreas(2022)}]{andreas-2022-language}
Jacob Andreas. 2022.
\newblock \href {https://aclanthology.org/2022.findings-emnlp.423} {Language
  models as agent models}.
\newblock In \emph{Findings of the Association for Computational Linguistics:
  EMNLP 2022}, pages 5769--5779, Abu Dhabi, United Arab Emirates. Association
  for Computational Linguistics.

\bibitem[{Attari et~al.(2019)Attari, Heckmann, and
  Schlangen}]{attari:sigdial19}
Nazia Attari, Martin Heckmann, and David Schlangen. 2019.
\newblock From explainability to explanation: Using a dialogue setting to
  elicit annotations with justifications.
\newblock In \emph{Proceedings of SIGdial 2019, Short Papers}, Stockholm,
  Sweden.

\bibitem[{Bara et~al.(2021)Bara, CH-Wang, and Chai}]{bara-etal-2021-mindcraft}
Cristian-Paul Bara, Sky CH-Wang, and Joyce Chai. 2021.
\newblock \href {https://aclanthology.org/2021.emnlp-main.85} {{M}ind{C}raft:
  Theory of mind modeling for situated dialogue in collaborative tasks}.
\newblock In \emph{Proceedings of the 2021 Conference on Empirical Methods in
  Natural Language Processing}, pages 1112--1125, Online and Punta Cana,
  Dominican Republic. Association for Computational Linguistics.

\bibitem[{Bartneck et~al.(2009)Bartneck, Kuli{\'{c}}, Croft, and
  Zoghbi}]{Bartneck2009}
Christoph Bartneck, Dana Kuli{\'{c}}, Elizabeth Croft, and Susana Zoghbi. 2009.
\newblock \href {https://doi.org/10.1007/s12369-008-0001-3} {{Measurement
  instruments for the anthropomorphism, animacy, likeability, perceived
  intelligence, and perceived safety of robots}}.
\newblock \emph{International Journal of Social Robotics}, 1(1):71--81.

\bibitem[{Bisk et~al.(2020)Bisk, Holtzman, Thomason, Andreas, Bengio, Chai,
  Lapata, Lazaridou, May, Nisnevich, Pinto, and Turian}]{Bisk2020}
Yonatan Bisk, Ari Holtzman, Jesse Thomason, Jacob Andreas, Yoshua Bengio, Joyce
  Chai, Mirella Lapata, Angeliki Lazaridou, Jonathan May, Aleksandr Nisnevich,
  Nicolas Pinto, and Joseph Turian. 2020.
\newblock \href {https://doi.org/10.18653/v1/2020.emnlp-main.703} {{Experience
  grounds language}}.
\newblock \emph{EMNLP 2020 - 2020 Conference on Empirical Methods in Natural
  Language Processing, Proceedings of the Conference}, pages 8718--8735.

\bibitem[{Bommasani et~al.(2021)Bommasani, Hudson, Adeli, Altman, Arora, von
  Arx, Bernstein, Bohg, Bosselut, Brunskill, Brynjolfsson, Buch, Card,
  Castellon, Chatterji, Chen, Creel, Davis, Demszky, Donahue, Doumbouya,
  Durmus, Ermon, Etchemendy, Ethayarajh, Fei-Fei, Finn, Gale, Gillespie, Goel,
  Goodman, Grossman, Guha, Hashimoto, Henderson, Hewitt, Ho, Hong, Hsu, Huang,
  Icard, Jain, Jurafsky, Kalluri, Karamcheti, Keeling, Khani, Khattab, Koh,
  Krass, Krishna, Kuditipudi, Kumar, Ladhak, Lee, Lee, Leskovec, Levent, Li,
  Li, Ma, Malik, Manning, Mirchandani, Mitchell, Munyikwa, Nair, Narayan,
  Narayanan, Newman, Nie, Niebles, Nilforoshan, Nyarko, Ogut, Orr,
  Papadimitriou, Park, Piech, Portelance, Potts, Raghunathan, Reich, Ren, Rong,
  Roohani, Ruiz, Ryan, R'e, Sadigh, Sagawa, Santhanam, Shih, Srinivasan,
  Tamkin, Taori, Thomas, Tram{\`e}r, Wang, Wang, Wu, Wu, Wu, Xie, Yasunaga,
  You, Zaharia, Zhang, Zhang, Zhang, Zhang, Zheng, Zhou, and
  Liang}]{Bommasani2021FoundationModels}
Rishi Bommasani, Drew~A. Hudson, Ehsan Adeli, Russ Altman, Simran Arora, Sydney
  von Arx, Michael~S. Bernstein, Jeannette Bohg, Antoine Bosselut, Emma
  Brunskill, Erik Brynjolfsson, S.~Buch, Dallas Card, Rodrigo Castellon,
  Niladri~S. Chatterji, Annie~S. Chen, Kathleen~A. Creel, Jared Davis, Dora
  Demszky, Chris Donahue, Moussa Doumbouya, Esin Durmus, Stefano Ermon, John
  Etchemendy, Kawin Ethayarajh, Li~Fei-Fei, Chelsea Finn, Trevor Gale,
  Lauren~E. Gillespie, Karan Goel, Noah~D. Goodman, Shelby Grossman, Neel Guha,
  Tatsunori Hashimoto, Peter Henderson, John Hewitt, Daniel~E. Ho, Jenny Hong,
  Kyle Hsu, Jing Huang, Thomas~F. Icard, Saahil Jain, Dan Jurafsky, Pratyusha
  Kalluri, Siddharth Karamcheti, Geoff Keeling, Fereshte Khani, O.~Khattab,
  Pang~Wei Koh, Mark~S. Krass, Ranjay Krishna, Rohith Kuditipudi, Ananya Kumar,
  Faisal Ladhak, Mina Lee, Tony Lee, Jure Leskovec, Isabelle Levent, Xiang~Lisa
  Li, Xuechen Li, Tengyu Ma, Ali Malik, Christopher~D. Manning, Suvir~P.
  Mirchandani, Eric Mitchell, Zanele Munyikwa, Suraj Nair, Avanika Narayan,
  Deepak Narayanan, Benjamin Newman, Allen Nie, Juan~Carlos Niebles, Hamed
  Nilforoshan, J.~F. Nyarko, Giray Ogut, Laurel Orr, Isabel Papadimitriou,
  Joon~Sung Park, Chris Piech, Eva Portelance, Christopher Potts, Aditi
  Raghunathan, Robert Reich, Hongyu Ren, Frieda Rong, Yusuf~H. Roohani, Camilo
  Ruiz, Jack Ryan, Christopher R'e, Dorsa Sadigh, Shiori Sagawa, Keshav
  Santhanam, Andy Shih, Krishna~Parasuram Srinivasan, Alex Tamkin, Rohan Taori,
  Armin~W. Thomas, Florian Tram{\`e}r, Rose~E. Wang, William Wang, Bohan Wu,
  Jiajun Wu, Yuhuai Wu, Sang~Michael Xie, Michihiro Yasunaga, Jiaxuan You,
  Matei~A. Zaharia, Michael Zhang, Tianyi Zhang, Xikun Zhang, Yuhui Zhang,
  Lucia Zheng, Kaitlyn Zhou, and Percy Liang. 2021.
\newblock \href {https://crfm.stanford.edu/assets/report.pdf} {On the
  opportunities and risks of foundation models}.
\newblock \emph{ArXiv}.

\bibitem[{Brennan(2000)}]{brennan:acl00}
Susan~E. Brennan. 2000.
\newblock Processes that shape conversation and their implications for
  computational linguistics.
\newblock In \emph{Proceedings of the 38th Annual Meeting of the Association
  for Computational Linguistics (ACL 2000)}, Hong Kong, China.

\bibitem[{Clark et~al.(2019)Clark, Lee, Chang, Kwiatkowski, Collins, and
  Toutanova}]{clark-etal-2019-boolq}
Christopher Clark, Kenton Lee, Ming-Wei Chang, Tom Kwiatkowski, Michael
  Collins, and Kristina Toutanova. 2019.
\newblock \href {https://doi.org/10.18653/v1/N19-1300} {{B}ool{Q}: Exploring
  the surprising difficulty of natural yes/no questions}.
\newblock In \emph{Proceedings of the 2019 Conference of the North {A}merican
  Chapter of the Association for Computational Linguistics: Human Language
  Technologies, Volume 1 (Long and Short Papers)}, pages 2924--2936,
  Minneapolis, Minnesota. Association for Computational Linguistics.

\bibitem[{Clark(1996)}]{clark:ul}
Herbert~H. Clark. 1996.
\newblock \emph{Using Language}.
\newblock Cambridge University Press, Cambridge.

\bibitem[{Das et~al.(2017)Das, Kottur, Moura, Lee, and
  Batra}]{dasetal:visdialrl}
Abhishek Das, Satwik Kottur, José M.~F. Moura, Stefan Lee, and Dhruv Batra.
  2017.
\newblock \href {https://doi.org/10.1109/ICCV.2017.321} {Learning cooperative
  visual dialog agents with deep reinforcement learning}.
\newblock In \emph{2017 IEEE International Conference on Computer Vision
  (ICCV)}, pages 2970--2979.

\bibitem[{de~Ruiter et~al.(2006)de~Ruiter, Mitterer, and
  Enfield}]{deruiteretal:corner}
J.P. de~Ruiter, H.~Mitterer, and N.J. Enfield. 2006.
\newblock Projecting the end of a speaker's turn: a cognitive cornerstone of
  conversation.
\newblock \emph{Language}, 82(3):504--524.

\bibitem[{Dingemanse et~al.(2023)Dingemanse, Liesenfeld, Rasenberg, Albert,
  Ameka, Birhane, Bolis, Cassell, Clift, Cuffari, De~Jaegher, Novaes, Enfield,
  Fusaroli, Gregoromichelaki, Hutchins, Konvalinka, Milton,
  Rączaszek-Leonardi, Reddy, Rossano, Schlangen, Seibt, Stokoe, Suchman,
  Vesper, Wheatley, and Wiltschko}]{beyond-single}
Mark Dingemanse, Andreas Liesenfeld, Marlou Rasenberg, Saul Albert, Felix~K.
  Ameka, Abeba Birhane, Dimitris Bolis, Justine Cassell, Rebecca Clift, Elena
  Cuffari, Hanne De~Jaegher, Catarina~Dutilh Novaes, N.~J. Enfield, Riccardo
  Fusaroli, Eleni Gregoromichelaki, Edwin Hutchins, Ivana Konvalinka, Damian
  Milton, Joanna Rączaszek-Leonardi, Vasudevi Reddy, Federico Rossano, David
  Schlangen, Johanna Seibt, Elizabeth Stokoe, Lucy Suchman, Cordula Vesper,
  Thalia Wheatley, and Martina Wiltschko. 2023.
\newblock \href {https://doi.org/https://doi.org/10.1111/cogs.13230} {Beyond
  single-mindedness: A figure-ground reversal for the cognitive sciences}.
\newblock \emph{Cognitive Science}, 47(1):e13230.

\bibitem[{Dreyfus(1981)}]{Dreyfus1981}
Hubert~L. Dreyfus. 1981.
\newblock From micro-worlds to knowledge: {AI} at an impasse.
\newblock In John Haugeland, editor, \emph{Mind Design}. MIT Press.

\bibitem[{Duan et~al.(2022)Duan, Yu, Tan, Zhu, and Tan}]{Duan2022}
Jiafei Duan, Samson Yu, Hui~Li Tan, Hongyuan Zhu, and Cheston Tan. 2022.
\newblock \href {https://doi.org/10.1109/TETCI.2022.3141105} {{A Survey of
  Embodied AI: From Simulators to Research Tasks}}.
\newblock \emph{IEEE Transactions on Emerging Topics in Computational
  Intelligence}, 6(2):230--244.

\bibitem[{Edlund et~al.(2008)Edlund, Gustafson, Heldner, and
  Hjalmarsson}]{edlundetal:humlike}
Jens Edlund, Joakim Gustafson, Mattias Heldner, and Anna Hjalmarsson. 2008.
\newblock Towards human-like spoken dialogue systems.
\newblock \emph{Speech Communication}, 50:630--645.

\bibitem[{Fern\'{a}ndez and Ginzburg(2002)}]{fernginz:corp}
Raquel Fern\'{a}ndez and Jonathan Ginzburg. 2002.
\newblock Non-sentential utterances in dialogue: A corpus-based study.
\newblock In \emph{Proceedings of the Third SIGdial Workshop on Discourse and
  Dialogue}, pages 15--26, Philadelphia, USA. ACL Special Interest Group on
  Dialog.

\bibitem[{Fern\'andez et~al.(2006)Fern\'andez, Lucht, Rodr\'iguez, and
  Schlangen}]{deawu:pttarub}
Raquel Fern\'andez, Tatjana Lucht, Kepa Rodr\'iguez, and David Schlangen. 2006.
\newblock \href
  {http://www.ling.uni-potsdam.de/~das/papers/fernandez\_etal\_slt2006.pdf}
  {Interaction in task-oriented human--human dialogue: The effects of different
  turn-taking policies}.
\newblock In \emph{Proceedings of the First International IEEE/ACL Workshop on
  Spoken Language Technology}, Palm Beach, Aruba.

\bibitem[{Flake and Fried(2020)}]{Flake2020}
Jessica~Kay Flake and Eiko~I. Fried. 2020.
\newblock \href {https://doi.org/10.1177/2515245920952393} {{Measurement
  Schmeasurement : Questionable Measurement Practices and How to Avoid Them}}.
\newblock \emph{Advances in Methods and Practices in Psychological Science},
  3(4):456--465.

\bibitem[{Frank et~al.(2023)Frank, Braginsky, Cachia, Coles, Hardwicke,
  Hawkins, Mathur, and Williams}]{franketal:experimentology}
Michael~C. Frank, Mika Braginsky, Julie Cachia, Nicholas Coles, Tom Hardwicke,
  Robert Hawkins, Maya Mathur, and Rondeline Williams. 2023.
\newblock \href {https://experimentology.io} {Experimentology: An open science
  approach to experimental psychology methods}.
\newblock Website.

\bibitem[{Fried et~al.(2022)Fried, Tomlin, Hu, Patel, and
  Nematzadeh}]{DBLP:journals/corr/abs-2211-08371}
Daniel Fried, Nicholas Tomlin, Jennifer Hu, Roma Patel, and Aida Nematzadeh.
  2022.
\newblock \href {https://doi.org/10.48550/arXiv.2211.08371} {Pragmatics in
  grounded language learning: Phenomena, tasks, and modeling approaches}.
\newblock \emph{CoRR}, abs/2211.08371.

\bibitem[{G{\"o}tze et~al.(2022)G{\"o}tze, {Paetzel-Pr{\"u}smann}, Liermann,
  Diekmann, and Schlangen}]{Goetze-2022-1}
Jana G{\"o}tze, Maike {Paetzel-Pr{\"u}smann}, Wencke Liermann, Tim Diekmann,
  and David Schlangen. 2022.
\newblock \href
  {http://www.lrec-conf.org/proceedings/lrec2022/pdf/2022.lrec-1.433.pdf} {The
  slurk interaction server framework: {{Better}} data for better dialog
  models}.
\newblock In \emph{Proceedings of the Language Resources and Evaluation
  Conference}, pages 4069--4078, {Marseille, France}. {European Language
  Resources Association}.

\bibitem[{Gu et~al.(2022)Gu, Stefani, Wu, Thomason, and
  Wang}]{gu-etal-2022-vision}
Jing Gu, Eliana Stefani, Qi~Wu, Jesse Thomason, and Xin Wang. 2022.
\newblock \href {https://doi.org/10.18653/v1/2022.acl-long.524}
  {Vision-and-language navigation: A survey of tasks, methods, and future
  directions}.
\newblock In \emph{Proceedings of the 60th Annual Meeting of the Association
  for Computational Linguistics (Volume 1: Long Papers)}, pages 7606--7623,
  Dublin, Ireland. Association for Computational Linguistics.

\bibitem[{Haugeland(1985)}]{Haugeland1985}
John Haugeland. 1985.
\newblock \emph{{Artificial Intelligence: The Very Idea}}.
\newblock MIT Press, Cambridge, Mass.

\bibitem[{Holler and Levinson(2019)}]{Holler2019}
Judith Holler and Stephen~C. Levinson. 2019.
\newblock \href {https://doi.org/10.1016/j.tics.2019.05.006} {{Multimodal
  Language Processing in Human Communication}}.
\newblock \emph{Trends in Cognitive Sciences}, pages 1--14.

\bibitem[{Howcroft et~al.(2020)Howcroft, Belz, Clinciu, Gkatzia, Hasan,
  Mahamood, Mille, van Miltenburg, Santhanam, and
  Rieser}]{howcroft-etal-2020-twenty}
David~M. Howcroft, Anya Belz, Miruna-Adriana Clinciu, Dimitra Gkatzia, Sadid~A.
  Hasan, Saad Mahamood, Simon Mille, Emiel van Miltenburg, Sashank Santhanam,
  and Verena Rieser. 2020.
\newblock \href {https://aclanthology.org/2020.inlg-1.23} {Twenty years of
  confusion in human evaluation: {NLG} needs evaluation sheets and standardised
  definitions}.
\newblock In \emph{Proceedings of the 13th International Conference on Natural
  Language Generation}, pages 169--182, Dublin, Ireland. Association for
  Computational Linguistics.

\bibitem[{Ilinykh et~al.(2019)Ilinykh, Zarrie{\ss}, and
  Schlangen}]{ilieatal:meetup19}
Nikolai Ilinykh, Sina Zarrie{\ss}, and David Schlangen. 2019.
\newblock Meetup! a corpus of joint activity dialogues in a visual environment.
\newblock In \emph{Proceedings of the 23rd Workshop on the Semantics and
  Pragmatics of Dialogue (SemDial 2019 / LondonLogue)}, London, UK.

\bibitem[{Ji et~al.(2022)Ji, Kojima, Rush, Suhr, Vong, Hawkins, and
  Artzi}]{ji-etal-2022-abstract}
Anya Ji, Noriyuki Kojima, Noah Rush, Alane Suhr, Wai~Keen Vong, Robert Hawkins,
  and Yoav Artzi. 2022.
\newblock \href {https://aclanthology.org/2022.emnlp-main.38} {Abstract visual
  reasoning with tangram shapes}.
\newblock In \emph{Proceedings of the 2022 Conference on Empirical Methods in
  Natural Language Processing}, pages 582--601, Abu Dhabi, United Arab
  Emirates. Association for Computational Linguistics.

\bibitem[{Kennington et~al.(2013)Kennington, Kousidis, and
  Schlangen}]{kenningtonetal:sigdial13}
Casey Kennington, Spyros Kousidis, and David Schlangen. 2013.
\newblock \href {http://www.aclweb.org/anthology/W/W13/W13-4030} {Interpreting
  situated dialogue utterances: an update model that uses speech, gaze, and
  gesture information}.
\newblock In \emph{Proceedings of the SIGDIAL 2013 Conference}, pages 173--182,
  Metz, France. Association for Computational Linguistics.

\bibitem[{Kocaballi et~al.(2019)Kocaballi, Laranjo, and
  Coiera}]{10.1093/iwc/iwz015}
Ahmet~Baki Kocaballi, Liliana Laranjo, and Enrico Coiera. 2019.
\newblock \href {https://doi.org/10.1093/iwc/iwz015} {{Understanding and
  Measuring User Experience in Conversational Interfaces}}.
\newblock \emph{Interacting with Computers}, 31(2):192--207.

\bibitem[{Kousidis et~al.(2012)Kousidis, Pfeiffer, Malisz, Wagner, and
  Schlangen}]{kousidis:mintlab}
Spyros Kousidis, Thies Pfeiffer, Zofia Malisz, Petra Wagner, and David
  Schlangen. 2012.
\newblock Evaluating a minimally invasive laboratory architecture for recording
  multimodal conversational data.
\newblock In \emph{Proceedings of the Interdisciplinary Workshop on Feedback
  Behaviors in Dialog at Interspeech 2012}, pages 39--42, Stevenson, WA, USA.

\bibitem[{Krauss and Weinheimer(1964)}]{krausswein:1964}
Robert~M. Krauss and Sidney Weinheimer. 1964.
\newblock Changes in reference phrases as a function of frequency of usage in
  social interaction: A preliminary study.
\newblock \emph{Psychonomic Science}, 1:266--278.

\bibitem[{Lapuschkin et~al.(2019)Lapuschkin, W{\"{a}}ldchen, Binder, Montavon,
  Samek, and M{\"{u}}ller}]{Lapuschkin2019}
Sebastian Lapuschkin, Stephan W{\"{a}}ldchen, Alexander Binder, Gr{\'{e}}goire
  Montavon, Wojciech Samek, and Klaus~Robert M{\"{u}}ller. 2019.
\newblock \href {https://doi.org/10.1038/s41467-019-08987-4} {{Unmasking Clever
  Hans predictors and assessing what machines really learn}}.
\newblock \emph{Nature Communications}, 10(1):1--8.

\bibitem[{Lazaridou et~al.(2017)Lazaridou, Peysakhovich, and
  Baroni}]{DBLP:conf/iclr/LazaridouPB17}
Angeliki Lazaridou, Alexander Peysakhovich, and Marco Baroni. 2017.
\newblock \href {https://openreview.net/forum?id=Hk8N3Sclg} {Multi-agent
  cooperation and the emergence of (natural) language}.
\newblock In \emph{5th International Conference on Learning Representations,
  {ICLR} 2017, Toulon, France, April 24-26, 2017, Conference Track
  Proceedings}. OpenReview.net.

\bibitem[{Levesque(2014)}]{Levesque2014}
Hector~J. Levesque. 2014.
\newblock \href {https://doi.org/10.1016/j.artint.2014.03.007} {{On our best
  behaviour}}.
\newblock \emph{Artificial Intelligence}, 212(1):27--35.

\bibitem[{Levinson(1979)}]{Levinson1979}
Stephen~C. Levinson. 1979.
\newblock {Activity types and language}.
\newblock \emph{Linguistics}, 17:365--399.

\bibitem[{Liang et~al.(2022)Liang, Bommasani, Lee, Tsipras, Soylu, Yasunaga,
  Zhang, Narayanan, Wu, Kumar, Newman, Yuan, Yan, Zhang, Cosgrove, Manning,
  R{\'{e}}, Acosta{-}Navas, Hudson, Zelikman, Durmus, Ladhak, Rong, Ren, Yao,
  Wang, Santhanam, Orr, Zheng, Y{\"{u}}ksekg{\"{o}}n{\"{u}}l, Suzgun, Kim,
  Guha, Chatterji, Khattab, Henderson, Huang, Chi, Xie, Santurkar, Ganguli,
  Hashimoto, Icard, Zhang, Chaudhary, Wang, Li, Mai, Zhang, and
  Koreeda}]{helm2022}
Percy Liang, Rishi Bommasani, Tony Lee, Dimitris Tsipras, Dilara Soylu,
  Michihiro Yasunaga, Yian Zhang, Deepak Narayanan, Yuhuai Wu, Ananya Kumar,
  Benjamin Newman, Binhang Yuan, Bobby Yan, Ce~Zhang, Christian Cosgrove,
  Christopher~D. Manning, Christopher R{\'{e}}, Diana Acosta{-}Navas, Drew~A.
  Hudson, Eric Zelikman, Esin Durmus, Faisal Ladhak, Frieda Rong, Hongyu Ren,
  Huaxiu Yao, Jue Wang, Keshav Santhanam, Laurel~J. Orr, Lucia Zheng, Mert
  Y{\"{u}}ksekg{\"{o}}n{\"{u}}l, Mirac Suzgun, Nathan Kim, Neel Guha,
  Niladri~S. Chatterji, Omar Khattab, Peter Henderson, Qian Huang, Ryan Chi,
  Sang~Michael Xie, Shibani Santurkar, Surya Ganguli, Tatsunori Hashimoto,
  Thomas Icard, Tianyi Zhang, Vishrav Chaudhary, William Wang, Xuechen Li,
  Yifan Mai, Yuhui Zhang, and Yuta Koreeda. 2022.
\newblock \href {https://doi.org/10.48550/arXiv.2211.09110} {Holistic
  evaluation of language models}.
\newblock \emph{CoRR}, abs/2211.09110.

\bibitem[{Liu et~al.(2016)Liu, Lowe, Serban, Noseworthy, Charlin, and
  Pineau}]{liu-etal-2016-evaluate}
Chia-Wei Liu, Ryan Lowe, Iulian Serban, Mike Noseworthy, Laurent Charlin, and
  Joelle Pineau. 2016.
\newblock \href {https://doi.org/10.18653/v1/D16-1230} {How {NOT} to evaluate
  your dialogue system: An empirical study of unsupervised evaluation metrics
  for dialogue response generation}.
\newblock In \emph{Proceedings of the 2016 Conference on Empirical Methods in
  Natural Language Processing}, pages 2122--2132, Austin, Texas. Association
  for Computational Linguistics.

\bibitem[{Madureira and Schlangen(2022)}]{Madureira-2022}
Brielen Madureira and David Schlangen. 2022.
\newblock \href {https://doi.org/10.18653/v1/2022.acl-short.73} {Can visual
  dialogue models do scorekeeping? exploring how dialogue representations
  incrementally encode shared knowledge}.
\newblock In \emph{Proceedings of the 60th Annual Meeting of the Association
  for Computational Linguistics (Volume 2: Short Papers)}, pages 651--664,
  Dublin, Ireland. Association for Computational Linguistics.

\bibitem[{Marr(1982)}]{Marr:Vision}
David Marr. 1982.
\newblock \emph{Vision: A Computational Investigation into the Human
  Representation and Processing of Visual Information}.
\newblock W.H. Freeman, San Francisco, USA.

\bibitem[{Minsky and Papert(1972)}]{MinskyPapert1972}
Marvin Minsky and Seymour Papert. 1972.
\newblock {Progress Report on Artificial intelligence}.
\newblock Technical report, MIT Artificial Intelligence Laboratory, Cambridge,
  Mass., USA.

\bibitem[{Mnih et~al.(2013)Mnih, Kavukcuoglu, Silver, Graves, Antonoglou,
  Wierstra, and Riedmiller}]{DBLP:journals/corr/MnihKSGAWR13}
Volodymyr Mnih, Koray Kavukcuoglu, David Silver, Alex Graves, Ioannis
  Antonoglou, Daan Wierstra, and Martin~A. Riedmiller. 2013.
\newblock \href {http://arxiv.org/abs/1312.5602} {Playing atari with deep
  reinforcement learning}.
\newblock \emph{CoRR}, abs/1312.5602.

\bibitem[{Padmakumar et~al.(2022)Padmakumar, Thomason, Shrivastava, Lange,
  Narayan-Chen, Gella, Piramuthu, Tur, and Hakkani-Tur}]{Padmakumar2022}
Aishwarya Padmakumar, Jesse Thomason, Ayush Shrivastava, Patrick Lange, Anjali
  Narayan-Chen, Spandana Gella, Robinson Piramuthu, Gokhan Tur, and Dilek
  Hakkani-Tur. 2022.
\newblock {TEACh: Task-Driven Embodied Agents That Chat}.
\newblock In \emph{Proceedings of the AAAI Conference on Artificial
  Intelligence}, pages 2017--2025.

\bibitem[{Raji et~al.(2021)Raji, Denton, Bender, Hanna, and
  Paullada}]{Raji-et-al-everything}
Deborah Raji, Emily Denton, Emily~M. Bender, Alex Hanna, and Amandalynne
  Paullada. 2021.
\newblock Ai and the everything in the whole wide world benchmark.
\newblock In \emph{Proceedings of the Neural Information Processing Systems
  Track on Datasets and Benchmarks}, volume~1. Curran.

\bibitem[{Ryle(1949)}]{Ryle:Concept}
Gilbert Ryle. 1949.
\newblock \emph{The Concept of Mind}.
\newblock Hutchinson \& Co.

\bibitem[{Schlangen(2019{\natexlab{a}})}]{Schlangen-2019-2}
David Schlangen. 2019{\natexlab{a}}.
\newblock \href {http://arxiv.org/abs/1908.11279} {Grounded agreement games:
  Emphasizing conversational grounding in visual dialogue settings}.
\newblock \emph{CoRR}, abs/1908.11279.

\bibitem[{Schlangen(2019{\natexlab{b}})}]{schlangen:tasks}
David Schlangen. 2019{\natexlab{b}}.
\newblock \href {http://arxiv.org/abs/1908.10747} {Language tasks and language
  games: On methodology in current natural language processing research}.
\newblock \emph{CoRR}, abs/1908.10747.

\bibitem[{Schlangen(2021)}]{schlangen-2021-targeting}
David Schlangen. 2021.
\newblock \href {https://doi.org/10.18653/v1/2021.acl-short.85} {Targeting the
  benchmark: On methodology in current natural language processing research}.
\newblock In \emph{Proceedings of the 59th Annual Meeting of the Association
  for Computational Linguistics and the 11th International Joint Conference on
  Natural Language Processing (Volume 2: Short Papers)}, pages 670--674,
  Online. Association for Computational Linguistics.

\bibitem[{Schlangen(2023)}]{Schlangen-2023}
David Schlangen. 2023.
\newblock \href {https://doi.org/10.48550/arXiv.2302.08590} {{What A Situated
  Language-Using Agent Must be Able to Do: {A} Top-Down Analysis}}.
\newblock \emph{CoRR}, abs/2302.08590.

\bibitem[{Schlangen et~al.(2018)Schlangen, Diekmann, Ilinykh, and
  Zarrieß}]{slurk.semdial18}
David Schlangen, Tim Diekmann, Nikolai Ilinykh, and Sina Zarrieß. 2018.
\newblock {slurk – A Lightweight Interaction Server For Dialogue Experiments
  and Data Collection}.
\newblock In \emph{Short Paper Proceedings of the 22nd Workshop on the
  Semantics and Pragmatics of Dialogue (AixDial / semdial 2018)}.

\bibitem[{Schlangen and Fern\'andez(2007)}]{schlandez:is07}
David Schlangen and Raquel Fern\'andez. 2007.
\newblock Speaking through a noisy channel - experiments on inducing
  clarification behaviour in human-human dialogue.
\newblock In \emph{Proceedings of Interspeech 2007}, Antwerp, Belgium.

\bibitem[{Schlangen and Lascarides(2002)}]{schlasc:edilog}
David Schlangen and Alex Lascarides. 2002.
\newblock \href {http://www.ling.uni-potsdam.de/~das/papers/edilog.pdf}
  {Resolving fragments using discourse information}.
\newblock In \emph{Proceedings of the 6th International Workshop on Formal
  Semantics and Pragmatics of Dialogue (EDILOG 2002)}, pages 161--168,
  Edinburgh.

\bibitem[{Sireci and Sukin(2013)}]{Sireci2013}
Stephen~G. Sireci and Tia Sukin. 2013.
\newblock {Test Validity}.
\newblock In K.~F. Geisinger, editor, \emph{APA Handbook of Testing and
  Assessment in Psychology: Vol. 1. Test Theory and Testing and Assessment in
  Industrial and Organizational Psychology}, chapter~4. The American
  Psychological Association.

\bibitem[{Srivastava et~al.(2022)Srivastava, Rastogi, Rao, Shoeb, Abid, Fisch,
  Brown, Santoro, Gupta, Garriga{-}Alonso, Kluska, Lewkowycz, Agarwal, Power,
  Ray, Warstadt, Kocurek, Safaya, Tazarv, Xiang, Parrish, Nie, Hussain, Askell,
  Dsouza, Rahane, Iyer, Andreassen, Santilli, Stuhlm{\"{u}}ller, Dai, La,
  Lampinen, Zou, Jiang, Chen, Vuong, Gupta, Gottardi, Norelli, Venkatesh,
  Gholamidavoodi, Tabassum, Menezes, Kirubarajan, Mullokandov, Sabharwal,
  Herrick, Efrat, Erdem, Karakas, and et~al.}]{bigbench2022}
Aarohi Srivastava, Abhinav Rastogi, Abhishek Rao, Abu Awal~Md Shoeb, Abubakar
  Abid, Adam Fisch, Adam~R. Brown, Adam Santoro, Aditya Gupta, Adri{\`{a}}
  Garriga{-}Alonso, Agnieszka Kluska, Aitor Lewkowycz, Akshat Agarwal, Alethea
  Power, Alex Ray, Alex Warstadt, Alexander~W. Kocurek, Ali Safaya, Ali Tazarv,
  Alice Xiang, Alicia Parrish, Allen Nie, Aman Hussain, Amanda Askell, Amanda
  Dsouza, Ameet Rahane, Anantharaman~S. Iyer, Anders Andreassen, Andrea
  Santilli, Andreas Stuhlm{\"{u}}ller, Andrew~M. Dai, Andrew La, Andrew~K.
  Lampinen, Andy Zou, Angela Jiang, Angelica Chen, Anh Vuong, Animesh Gupta,
  Anna Gottardi, Antonio Norelli, Anu Venkatesh, Arash Gholamidavoodi, Arfa
  Tabassum, Arul Menezes, Arun Kirubarajan, Asher Mullokandov, Ashish
  Sabharwal, Austin Herrick, Avia Efrat, Aykut Erdem, Ayla Karakas, and et~al.
  2022.
\newblock \href {https://doi.org/10.48550/arXiv.2206.04615} {Beyond the
  imitation game: Quantifying and extrapolating the capabilities of language
  models}.
\newblock \emph{CoRR}, abs/2206.04615.

\bibitem[{Suits(1978)}]{suits:grasshopper}
Bernard Suits. 1978.
\newblock \emph{The Grasshopper: Games, Life, and Utopia}.
\newblock The University of Toronto Press, Toronto, Canada.

\bibitem[{Sundar and Heck(2022)}]{sundar-heck-2022-multimodal}
Anirudh Sundar and Larry Heck. 2022.
\newblock \href {https://doi.org/10.18653/v1/2022.nlp4convai-1.12} {Multimodal
  conversational {AI}: A survey of datasets and approaches}.
\newblock In \emph{Proceedings of the 4th Workshop on NLP for Conversational
  AI}, pages 131--147, Dublin, Ireland. Association for Computational
  Linguistics.

\bibitem[{Turian et~al.(2003)Turian, Shen, and
  Melamed}]{turian-etal-2003-evaluation}
Joseph~P. Turian, Luke Shen, and I.~Dan Melamed. 2003.
\newblock \href {https://aclanthology.org/2003.mtsummit-papers.51} {Evaluation
  of machine translation and its evaluation}.
\newblock In \emph{Proceedings of Machine Translation Summit IX: Papers}, New
  Orleans, USA.

\bibitem[{Turing(1950)}]{turing:test}
Alan Turing. 1950.
\newblock Computing machinery and intelligence.
\newblock \emph{Mind}, 59:433--460.

\bibitem[{von Ahn and Dabbish(2004)}]{vonAhn:espgame}
Luis von Ahn and Laura Dabbish. 2004.
\newblock \href {https://doi.org/10.1145/985692.985733} {Labeling images with a
  computer game}.
\newblock In \emph{Proceedings of the SIGCHI Conference on Human Factors in
  Computing Systems}, CHI '04, pages 319--326, New York, NY, USA. ACM.

\bibitem[{Walker et~al.(1998)Walker, Litman, Kamm, and
  Abella}]{walkeretal:paradise}
Marilyn~A. Walker, Diane~J. Litman, Candace~A. Kamm, and Alicia Abella. 1998.
\newblock Evaluating spoken dialogue agents with {PARADISE}: Two case studies.
\newblock \emph{Computer Speech and Language}, 12(3).

\bibitem[{Wang et~al.(2019{\natexlab{a}})Wang, Pruksachatkun, Nangia, Singh,
  Michael, Hill, Levy, and Bowman}]{superGLUE}
Alex Wang, Yada Pruksachatkun, Nikita Nangia, Amanpreet Singh, Julian Michael,
  Felix Hill, Omer Levy, and Samuel~R. Bowman. 2019{\natexlab{a}}.
\newblock \href {http://arxiv.org/abs/1905.00537} {{SuperGLUE: A Stickier
  Benchmark for General-Purpose Language Understanding Systems}}.
\newblock In \emph{NeurIPS}, July, pages 1--30.

\bibitem[{Wang et~al.(2019{\natexlab{b}})Wang, Singh, Michael, Hill, Levy, and
  Bowman}]{Wang2019}
Alex Wang, Amanpreet Singh, Julian Michael, Felix Hill, Omer Levy, and
  Samuel~R. Bowman. 2019{\natexlab{b}}.
\newblock \href {http://arxiv.org/abs/1804.07461} {{GLUE: A Multi-Task
  Benchmark and Analysis Platform for Natural Language Understanding}}.
\newblock In \emph{ICLR 2019}, pages 1--20.

\bibitem[{Winograd(1972)}]{winograd:shrdlu}
Terry Winograd. 1972.
\newblock Understanding natural language.
\newblock \emph{Cognitive Psychology}, 3(1):1--191.

\bibitem[{Wittgenstein(1984 {[1953]})}]{Witt:PU-corr}
Ludwig Wittgenstein. 1984 {[1953]}.
\newblock \emph{Tractatus Logicus Philosophicus und Philosophische
  Untersuchungen}, volume~1 of \emph{Werkausgabe}.
\newblock Suhrkamp, Frankfurt am Main.

\bibitem[{Zarrie{\ss} et~al.(2016)Zarrie{\ss}, Hough, Kennington,
  Manuvinakurike, DeVault, Fern{\'a}ndez, and Schlangen}]{pentoref2016}
Sina Zarrie{\ss}, Julian Hough, Casey Kennington, Rames Manuvinakurike, David
  DeVault, Raquel Fern{\'a}ndez, and David Schlangen. 2016.
\newblock Pentoref: A corpus of spoken references in task-oriented dialogues.
\newblock In \emph{Proceedings of LREC 2016}, Portoroz, Slovenia.

\end{thebibliography}
\bibliographystyle{acl_natbib}

\appendix

\section{Example Appendix}
\label{sec:A}

Table~\ref{tab:class} shows the full taxonomy, with explanations of all attributes and values.

\begin{table*}[!ht]
\small
    \centering
    \begin{tabular}{|l|p{.2\linewidth}|p{.4\linewidth}|}
    \hline
        \textbf{Category / Feature} & \textbf{Possible Values} & \textbf{Description} \\ \hline
      \hline
        \textbf{Environment} & ~ & Characterises the relevant objects and configurations, and how they are presented to players \\ \hline
        presence & ready-to-hand / absent & are players talking about objects that are immediately perceivable to them or not? \\ \hline
        object familiarity & instance familiar / type familiar / unfamiliar & prior to game, do they know object instance (Barack Obama), type (fridge), or likely not at all (pento pieces) \\ \hline
        scene familiarity & instance familiar / type familiar / unfamiliar & same for constellations of objects \\ \hline
      access & immediate / mediated & are objects physically present or via interface? \\ \hline
      reality & real / virtual & are objects real or computer represented? \\ \hline
        fidelity & high / low & realism of representation \\ \hline
        dynamics & continuous / discrete / static & how changes of the environment proceed \\ \hline
        env action space & object manipulation / viewpoint man. / none & what can be done to environment \\ \hline
      \hline
        \textbf{Setting} & ~ & Characterises how the players can interact with each other \\ \hline
        verbal action channel & spoken / written & w/ variations on turn taking, e.g. "free turn taking", "push to talk", "turn-based", "round-robin" \\ \hline
        mutual observability & real / avatar / none & whether other player is visible / embodied \\ \hline
        task action channel & in-environment / symbolic feedback / none & how actions of other player are perceived (symb feedback would be e.g. just info whether they picked the right object, w/o player seeing the picking action) \\ \hline
        task common ground & single game / repeated games & whether players (knowingly) play repeated rounds \\ \hline
        env. common ground & full / partial / none & whether players are in same environment or not \\ \hline
      \hline
        \textbf{Game} & ~ & Characterises the goal of the interaction and the constraints on how to reach it \\ \hline
        role equality & equal / specialised / sequentially-equal & do players have the same action space or not (e.g., instruction giver / follower); "sequentially-equal" meaning they swap roles \\ \hline
        verbal action space & unrestricted / restricted & whether they can talk freely or are limited to range of utterances (e.g., just "yes" or "no"); by player \\ \hline
        goal information & symmetric / asym. / complementary / none & whether one player has solution, or both, or both have (different) parts; none means neither player knows more than general goal spec (e.g., like in chess) \\ \hline
        goal type & (games can contain several goal types) & ~ \\ \hline
        ~ & reference  & identify object(s) \\ \hline
        ~ & information & request / provide information \\ \hline
        ~ & construction & configure objects \\ \hline
        ~ & navigation & go somewhere / direct somewhere \\ \hline
        ~ & negotiation & agree on something \\ \hline
        ~ & teaching & teach / learn something \\ \hline
        scoring & binary / graded / none & measure of immediate task success (Interaction as a whole might additionally be evaluated otherwise as well) \\ \hline
        score impact & yes / no & whether players are motivated to achieve good score \\ \hline
        anticipated strategy & cooperative / collaborative / anti-collaborative & by player; whether player is expected to facilitate other player's goals, or has own goals which coincide (or not) w/ other player and for which other player is needed \\ \hline
    \end{tabular}
    \caption{The proposed fine-grained classification scheme}
    \label{tab:class}
  \end{table*}

\end{document}